\pdfoutput=1
\documentclass[11pt]{article}

\usepackage[preprint]{coling}

\usepackage{times}
\usepackage{latexsym}

\usepackage[T1]{fontenc}
\usepackage{xcolor}
\usepackage[utf8]{inputenc}
\usepackage{multirow}
\usepackage{microtype}
\usepackage{inconsolata}
\usepackage{arydshln}
\usepackage{adjustbox}
\usepackage{booktabs}
\usepackage{graphicx}

\usepackage{amsmath}
\title{TOOL-ED: Enhancing Empathetic Response Generation with the Tool Calling Capability of LLM}

\author{
 Huiying~Cao, Yiqun~Zhang, Shi~Feng$^{\dag}$, Xiaocui~Yang, Daling~Wang, Yifei~Zhang\hspace{0.4em} \\
 School of Computer Science and Engineering, Northeastern University, Shenyang, China \\
 \texttt{cao\_huiy@163.com}\\
 \texttt{\{zhangyiqun344, neu.yangxiaocui\}@gmail.com}\\ 
\texttt{\{fengshi, wangdaling, zhangyifei\}@cse.neu.edu.cn}
}
\begin{document}
\maketitle
\begin{abstract}

\def\thefootnote{\dag}\footnotetext{Corresponding author.}\def\thefootnote{\arabic{footnote}}
Empathetic conversation is a crucial characteristic in daily conversations between individuals.
Nowadays, Large Language models (LLMs) have shown outstanding performance in generating empathetic responses. 
Knowledge bases like COMET can assist LLMs in mitigating illusions and enhancing the understanding of users' intentions and emotions.
However, models remain heavily reliant on fixed knowledge bases and unrestricted incorporation of external knowledge can introduce noise.
Tool learning is a flexible end-to-end approach that assists LLMs in handling complex problems.
In this paper, we propose \textbf{E}motional \textbf{K}nowledge \textbf{T}ool \textbf{C}alling (EKTC) framework, which encapsulates the commonsense knowledge bases as empathetic tools, enabling LLMs to integrate external knowledge flexibly through tool calling.
In order to adapt the models to the new task, we construct a novel dataset TOOL-ED based on the E{\small MPATHETIC}{\small MPATHETIC}D{\small IALOGUE} (ED) dataset.
We validate EKTC on the ED dataset, and the experimental results demonstrate that our framework can enhance the ability of LLMs to generate empathetic responses effectively.
Our code is available at { \url{https://github.com/caohy123/EKTC}}
\end{abstract}

\begin{figure}[h]
\centering
\includegraphics[width=0.5\textwidth]{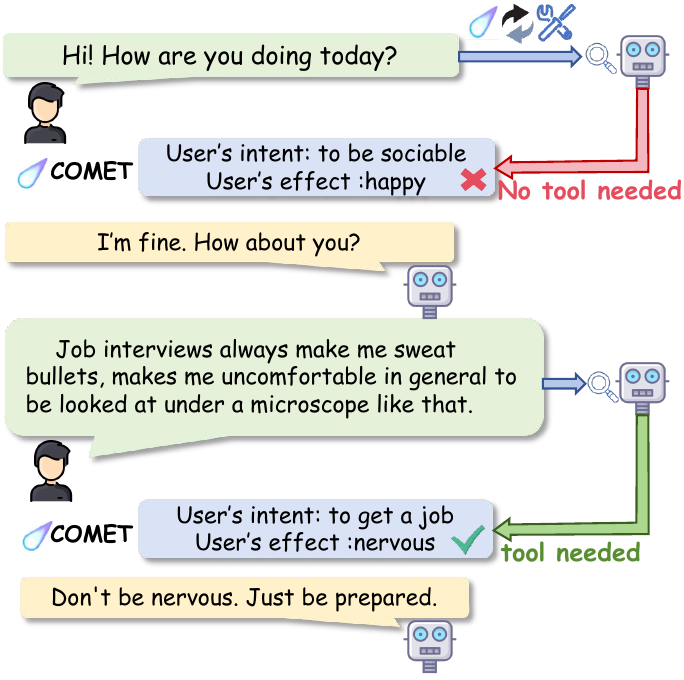}
\caption{Architecture for the application of a specified tool in an example of empathetic dialogue. The model owns the ability to determine the timing of empathetic tools calling actively.}

\label{figure_1}
\end{figure}

\section{Introduction}
As a hot topic in building humanlike chatbots, empathetic dialogue aims to enhance the ability to fully understand the users' emotions and make appropriate responses, which plays an essential role in establishing and maintaining harmonious social connections \cite{keskin2014isn, wang2023enhancing}.
There are several works focused on enhancing the empathetic ability of models by understanding dialogue context from the perspective of emotions and sentiment cognition \cite{lin-etal-2019-moel, majumder-etal-2020-mime, li-etal-2020-empdg, yang2024iterative}.
Due to the complexity of conversations, dialogues often contain implicit knowledge \cite{zhou-etal-2023-case, zhao-etal-2023-dont}, including psychological states, potential causality and so on. Many researchers have infused external knowledge to assist models in understanding more detailed information and generating more comprehensive responses \cite{sabour2022cem, li2022knowledge}.
Nowadays, Large Language Models (LLMs) have shown excellent comprehension and powerful generation abilities by intermediate thinking or commonsense reasoning without fine-tuning \cite{wang2023chain, 10.5555/3495724.3495883, liu2024chatzero, chae2023dialogue, wei2022chain}.
The adoption of knowledge base allows LLMs to retrieve specific information such as users' intent and emotions, thereby enhancing their empathetic capabilities and effectively mitigating hallucination phenomena during the response generation process \cite{yang2024enhancing, qian2023harnessing}.
However, some utterances such as the daily greeting statement "Hi" do not require external knowledge base assistance for analysis.
Accordingly, the overly strong reliance on a specific knowledge base substantially reduces the flexibility of the models, and the continuous influx of external knowledge introduces additional noise into the model.

\begin{figure*}
    \centering
    \includegraphics[width=\textwidth]{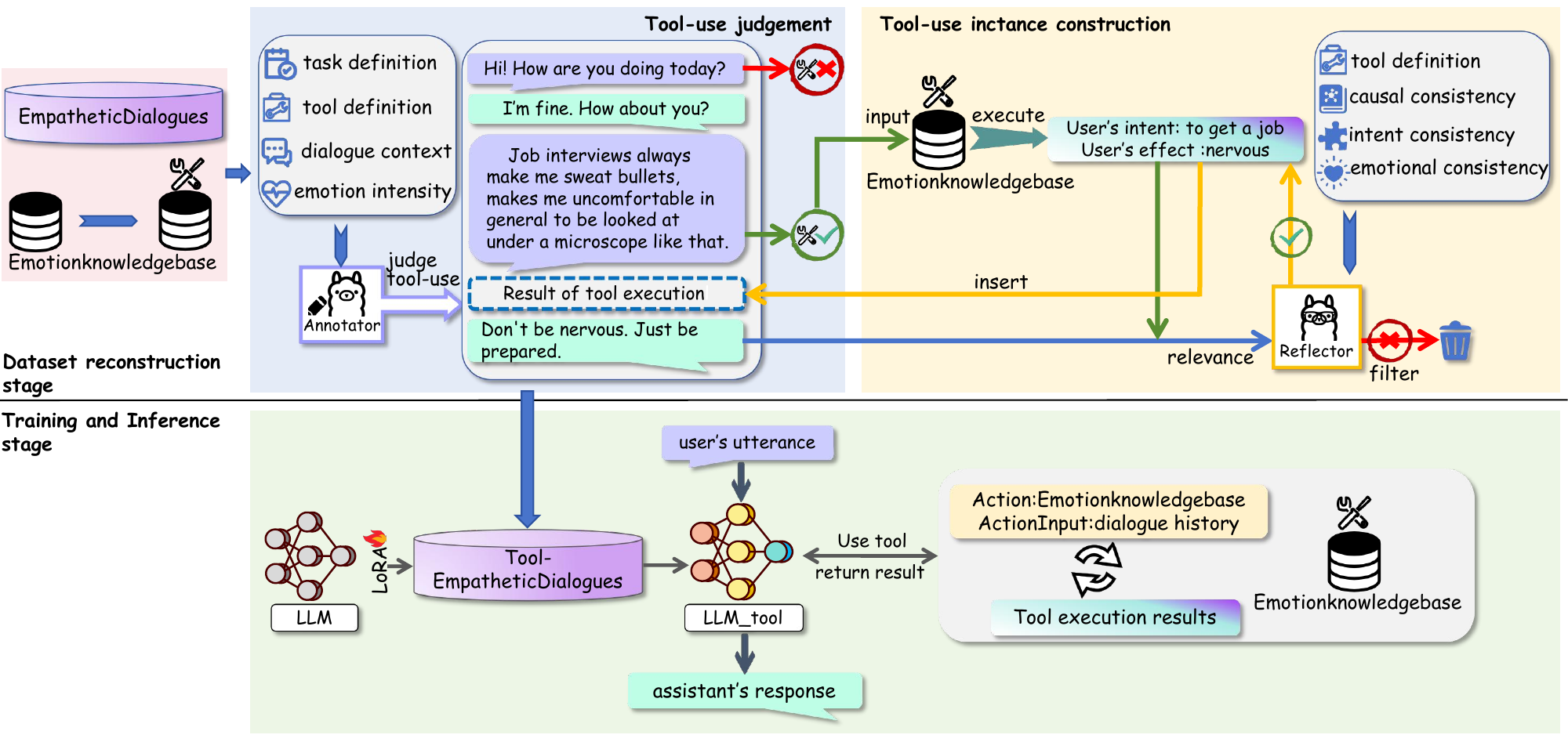}
    \caption{The architecture of the EKTC framework consists of two stages: Dataset Reconstruction \& Training and Inference stage. In the Dataset Reconstruction stage, the commonsense knowledge base is defined as the $Emotionknowledgebase$ tool. Annotator determines the tool calls based on the context and sends the corresponding content to the tool. Reflector judges the relevance between the execution results and the golden response in the dataset, inserting the highly relevant result into the constructed dataset. In the Training and Inference stage, the LLMs are trained on the constructed dataset for active invocation during inference.}
    \label{figure_2}
\end{figure*}

We prospect to enable LLMs to autonomously access external knowledge base in the empathetic conversation, rather than employing it in each round of dialogue. 
% Tool learning significantly improves the ability of LLMs to solve complex problems off own bat \cite{wang2024empowering}.
% LLMs can accquire and integrate external knowledge dynamically in the end-to-end way \cite{qu2024tool, zhao2024let, ji2023survey, zhang2023siren}.
Tool learning is a promising approach that enables LLMs to dynamically acquire external knowledge, thereby enhancing their ability to independently solve complex problems \cite{qu2024tool, zhao2024let, ji2023survey, zhang2023siren}.
% Additionally, if the model owns the ability to utilize the tool after training, the tools can be swapped out independently of the model as plugins.
Tools can be swapped out independently of LLMs as plugins if the model owns the ability to utilize the tool.
So we define the commonsense knowledge bases as tools, allowing LLMs to flexibly introduce external knowledge based on the dialogue context. 
% Although different commonsense knowledge bases may not have significant relevance, they can provide external knowledge to the model in an end-to-end manner once defined as tools.
% However, most tool learning tasks are based on the explicit queries from the users. In contrast, as for the empathetic dialogue task, there is no precise instruction from the users, which requires the model to actively decide whether to utilize the tools based on various factors such as the user's emotional intensity and the context of the dialogue.
However, in contrast with previous work on tool-use instances, the task on empathetic response generation does not involve direct inquiries or explicit requests from users for addressing specific issues \cite{qin2023toolllm, gao2024confucius, tang2023toolalpaca}. Instead, models require to decide whether to employ the empathetic tool based on a comprehensive assessment of the user's emotional intensity and contextual factors actively.

% Although different commonsense knowledge bases may not have significant relevance, they can provide external knowledge to the model in an end-to-end manner once defined as tools.
% However, previous work on automatic tool-use instances generation typically relies on the explicit queries user provided to invoke tools for solving specific problems. 
% In contrast, as for the empathetic dialogue task, users do not give explicit instructions, which requires the model to actively decide whether to utilize the tools based on various factors such as the user's emotional intensity and the context of the dialogue.

Therefore, we propose \textbf{E}motional \textbf{K}nowledge \textbf{T}ool \textbf{C}alling (EKTC) framework, which is designed to automatically generate multi-turn tool-use instance for empathetic responses generation, enabling LLMs to perform dynamic commonsense reasoning in an end-to-end manner with minimal human and material resources, as demonstrated in the Figure~\ref{figure_1}.
 % As shown in the Figure~\ref{figure_1}, we define the commonsense knowledge base as the empathetic tool, utilizing external information to stimulate the relevant knowledge encoded by LLM.
% In this way, LLM can utilize the commonsense knowledge base more effectively in a plug and play manner at the appropriate time.
Although different commonsense knowledge bases have little relevance, they can be easily switched once defined as tools.
So we opt COMET \cite{bosselut-etal-2019-comet} as a representative tool and insert tool-use trajectory into the E{\small MPATHETIC}{\small MPATHETIC}D{\small IALOGUE} (ED) dataset to construct an innovative dataset, TOOL-ED.
We aim to enable the model to acquire the ability to use empathetic tools, by fine-tuning them on TOOL-ED.
% To enable the model to discern the appropriate timing for invoking the knowledge base tool, we construct TOOL-ED corpus, defining COMET \cite{bosselut2019comet} as a representative tool and inserting tool usage trajectory into the ED dataset.
% The EKTC-based models employ empathetic tools at appropriate time based on the emotional intensity and various factors in the context in a plug-and-play manner.
Extensive testing experiments on the ED dataset \cite{rashkin2018towards} illustrate that our paradigm is able to take advantage of external tool, efficiently improving the quality of empathic response generation.
Our contributions can be summarized as follows:

(i) We propose a novel framework EKTC for empathetic dialogue. To our best knowledge, we are the first to use the tool learning paradigm to enhance empathetic abilities of LLMs.

(ii) We reconstruct a new dataset called TOOL-ED based on the ED dataset with the assistance of LLMs, which can be served as a benchmark for simulating the use of empathetic tools.

(iii) We define two distinct knowledge bases as tools and validate the generalization of EKTC through a plug-and-play manner. We conduct extensive experiments and analysis on the ED dataset and results demonstrate the effectiveness of our approach.

% \begin{itemize}
% \item[$\bullet$] We propose a novel framework EKTC for empathetic dialogue. To our best knowledge, we are the first to use the tool learning paradigm to enhance empathetic abilities of LLMs
% \item[$\bullet$] We reconstruct a new dataset called TOOL-ED based on the ED dataset with the assistance of LLMs, which can be served as a benchmark for simulating the use of empathetic tools.
% \item[$\bullet$] We define two distinct knowledge bases as tools and validate the generalization of EKTC through a plug-and-play manner.
% We conduct extensive experiments and analysis on the ED dataset and results demonstrate the effectiveness of our approach.
% % to demonstrate the text generation quality of the EKTC-based models.
% % To further validate the generalization of our framework, we also test it employing other knowledge base as an alternative tool. 
% % The experimental results affirm the effectiveness of our approach.

% \end{itemize}

\section{Related Work}
\subsection{Empathetic Response Generation}
Empathetic response generation refers to the process of generating effective responses that resonate emotionally with them in a conversation by thoroughly understanding the user's viewpoints and emotional state.
% ED dataset \cite{rashkin2018towards} is an open-sourced benchmark dataset which is widely used for empathetic response generation.
Some previous works have constructed empathetic dialogue systems by categorizing emotions, applying emotional cues \cite{huang2024generating,  song2019generating}, and incorporating external information \cite{li-etal-2020-empdg, sabour2022cem, cai-etal-2023-improving, cai2024pecer}.
However, limited parameters of the models restrict their ability to capture and convey complex emotions.
% \cite{sabour2022cem} and \cite{li2022knowledge} respectively use pretrained language models and graph neural network structures to incorporate commonsense reasoning into the empathetic response generation task.
% Pre-trained language models and graph neural network structures can be incorporate commonsense reasoning into the empathetic response generation task \cite{sabour2022cem} and \cite{li2022knowledge}.
\citeauthor{sabour2022cem} and \citeauthor{li2022knowledge} implement pre-trained language models and graph neural network structures to introduce common sense reasoning into the core of the empathy task.
Nowadays, LLMs like ChatGPT \cite{achiam2023gpt} and LLaMA3 \cite{touvron2023llama} are pretrained on vast amounts of data, covering extensive knowledge \cite{chen2023llm, wang2023robustness, sun2023chatgpt}. 
Through techniques like instruction fine-tuning, the models excel in various tasks \cite{he2021towards, liu2024hift}. 
Some researchers enhance the models' empathetic abilities by leveraging context learning, building prompt templates, combining LLMs with small models and augmenting commonsense knowledge \cite{lee2022does, liu2024knowdt, yang2024enhancing, qian2023harnessing}.
% Nevertheless, LLMs still exhibit hallucination phenomena in understanding external knowledge and during dialogue context processing.
Nevertheless, the continuous introduction of external knowledge may also lead to noise effects.
\subsection{LLM Tool Learning}
Nowadays, effectively leveraging tools in conjunction with LLMs to solve complex problems has become an effective paradigm.
Tool learning can be classified into two categories \cite{tang2023toolalpaca}:
% the first is directly leveraging the powerful tool usage capabilities of LLMs to interact with various tools directly \cite{qin2023toolllm, hsieh2023tool, li-etal-2023-api, shi2024learning},
The first approach involves LLMs with strong tool-use capabilities interacting directly with external tools \cite{qin2023toolllm, hsieh2023tool, li-etal-2023-api, shi2024learning}, 
and the second approach involves fine-tuning models to use specific tools through supervised learning on a specialized dataset \cite{parisi2022talm, schick2024toolformer, qin2304tool, qiao2024autoact, yang2024gpt4tools}. 
% \citeauthor{lu2024chameleon} used LLM for multimodal question answering and developed various tools, such as other LLMs, visual models, web search, and Python functions.
\citeauthor{lu2024chameleon} combine LLMs with various tools, such as pre-trained visual models, web search engines, python functions and heristic-based modules.
The combination of LLMs with such tools extends their ability to handle tasks involving both textual and visual data, as well as real-time information retrieval through web searches and external APIs.
\citeauthor{tang2023toolalpaca} fine-tune the 7B and 13B parameter alpaca derivatives on a corpus composed of generated OpenAPI specifications and descriptions, allowing for multiple rounds of interaction. 
\citeauthor{gou2023tora} enables LLM to solve complex mathematical problems by reasoning and interacting with external computing tools.
This paper defines a new framework for empathetic dialogue processes based on tuning-based theory of tool learning and constructs scenarios for the use of the empathetic tools.

\section{Method}
\subsection{EKTC}
Continuously injecting external commonsense knowledge may lead to extra noise, so the timing of calling empathetic tools is crucial for the quality of responses. However, existing work lacks attention to the appropriate invocation of knowledge bases.
Our goal is to innovatively apply the principles of tool learning, enabling models to flexibly utilize knowledge bases in a plug-and-play manner and efficiently integrate external knowledge during the generation of empathetic responses.

The overview of EKTC framework is shown in Figure~\ref{figure_2}. 
We firstly define the commonsense knowledge base as an empathetic tool and regard the utterances from assistant in the ED dataset as the golden responses.
 To equip the model with the ability to autonomously invoke tools with minimal human and material resources, LLaMA3 \cite{llama3modelcard} functions as Annotator to determine the timing of tool invocation based on the comprehensive instructions we provide, while also serving as Reflector to select the tool invocation processes with high relevance to the golden responses.
Finally, the high-quality tool-use instance are inserted into the constructed dataset.
% After being fine-tuned on the reconstructed the TOOL-ED dataset, the LLM is capable of implementing tool calls and providing empathetic responses in a timely and efficient manner.
After being fine-tuned on the constructed dataset, the LLM can effectively implement tool call and empathetic responses.

\begin{figure}[h]
\centering
\includegraphics[width=0.5\textwidth]{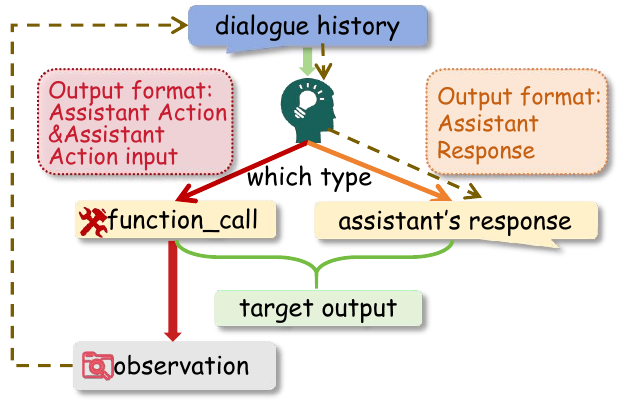}
\caption{Inference details of empathetic response generation task based on tool learning.}
\label{figure_3}
\end{figure}

The inference process of the fine-tuned model is illustrated in Figure~\ref{figure_3}. 
Following ReAct \cite{yao2022react}, we employ an (action, observation) format template to guide LLM in accomplishing the task.
Based on the dialogue history, the target output of the model can be divided into two main categories: 

\textbf{(i)} \textbf{Tool Call Responses} (function\_call)
When the model identifies to apply the empathetic tool, it will output "ASSISTANT Action" with the tool name and the corresponding prompt "ASSISTANT Action Input" with the arguments input into the tool. 
For example, the output (Action: $EmotionKnowledgebase$ Action Input: {"prompt": "I was surprised when my mom bought me a car"}) indicates that the model will input the user's utterance as the parameter to the $EmotionKnowledgebase$ tool.
Then the tool execution results will be incorporated to the dialogue history via data flow, served as observation.
After obtaining the dialogue history with observations, the fine-tuned model will proactively generate text responses.
In this way, the comnmonsense knowledge can be seamlessly integrated with the model.

\textbf{(ii)} \textbf{Original text Responses} (assistant)
% If the tool is determined for invocation, the execution results will be transmitted back to the model via data flow and added to the dialogue history.
If the model does not execute a tool call, it will generate the final textual empathetic response based on the dialogue history.
% Through multiple rounds of interaction, the model will provide a comprehensive empathetic response based on dialogue history.

\subsection{Tool Definition}
% LLMs store knowledge widely through their weights.
% To stimulate the abundant knowledge stored in the model during pre-training and minimize the impact of noise, 
In this paper, we define the commonsense knowledge base as the empathetic tool, called $EmotionKnowledgebase$, enabling the model to flexibly access external information in a plug and play manner.
In this way, the model automatically treat the dialogue context as arguments that are required to input into the empathetic tool.
Based on the given conversation context $C$, the execution process of the empathetic tool is as follows:

\begin{align}
        \begin{aligned}
        \textit{observation} \! = \! \textit{EmotionKnowledgebase}(C)
        \end{aligned}
\end{align}

\noindent where $observation$ is the result of the tool.
The following is the definition details of $Emotion\-$
$Knowledgebase$.

Referring to \cite{qian2023harnessing}, we utilize the COMET BART version \cite{hwang2021comet} trained on $ATOMIC_{20}^{20}$ as the empathetic tool ($EmotionKnowledgebase$), defining it as the API for generating commonsense inferences of five types of relations $( x_{\text{Content}}, x_{\text{Need}}, x_{\text{Want}}, x_{\text{Effect}}, x_{\text{React}} )$ based on the dialogue context.
Given the dialogue context $C$, the execution result of the tool is:
{
\begin{equation}
\begin{aligned}
&result\_r=COMET_{BART}(C, r)\\
&observation = \underset{R}{\oplus} result\_r
\end{aligned}
\end{equation}
}

\noindent where \text{\( R= \{ x_{\text{Content}}, x_{\text{Need}}, x_{\text{Want}}, x_{\text{Effect}}, x_{\text{React}} \} \)}, and \text{\( r \in R \)}.

% where \text{\( r \in \{ x_{\text{Content}}, x_{\text{Need}}, x_{\text{Want}}, x_{\text{Effect}}, x_{\text{React}} \} \)}.

In terms of knowledge generation models, CICERO \cite{shen2022multiview} has demonstrated outstanding abilities, especially in generating knowledge related to emotions. 
By utilizing the emotional response relationships, potential subsequent event, motivations, and causal relationships provided by CICERO, emotional recognition and conversational reasoning abilities of the model can be largely enhanced, improving the quality and effectiveness of interactions with users.
Therefore, we also define CICERO as an empathetic tool, in which the process of generating knowledge by the tool is as follows:
% {
% \begin{equation}
% observation=CICERO(C, r)
% \end{equation}
% }

% where \text{\( r \in \{ Cause, SubEv, Motiv, React \} \)}.

{
\begin{equation}
\begin{aligned}
&result\_r=CICERO(C, r)\\
&observation = \underset{R}{\oplus} result\_r
\end{aligned}
\end{equation}
}
% {
% \begin{equation}
% observation= \underset{R}{\oplus} result\_r
% \end{equation}
% }

\noindent where \text{\( R= \{ Cause, SubEv, Motiv, React \} \)}, and \text{\( r \in R \)}.

\subsection{Dataset Construction}
 % Vanilla models lack the capability to independently determine the optimal timing for invoking empathetic tools, so we need to create a dataset, TOOL-ED, specifically tailored for the tool Learning based empathetic response generation task to enable the model to adapt to such scenarios.

 In order to enable the vanilla model to independently determine the optimal timing for tool calls and effectively integrate external knowledge, we need to create a dataset, TOOL-ED, specifically tailored for the tool learning based empathetic response generation task.
 % for adapting to such scenarios.
% Therefore, we construct the TOOL-ED dataset based on the ED dataset.
% Due to the lack of precise instructions for addressing specific issues in the empathetic conversations, models need to acutely perceive the contextual situation during the dialogue and proactively invoke knowledge base tools to introduce external knowledge. However, models do not possess the capability to independently analyze the timing for invoking the empathetic tool, we need to create a dataset specifically tailored for this task to enable the model to adapt to such scenarios.

% First, we reformat the ED dataset into a chatbot format compatible with the ShareGPT\footnote{\href{https://sharegpt.com}{https://sharegpt.com}}.
% , including roles for $user$, $assistant$, $function\_call$, and $observation$.
% Then we add special tokens before each role's response, and insert an end-of-text marker to the end of each assistant output.
% During inference, the model is configured to stop after each round of responses.
% Since users do not provide explicit queries in empathetic dialogue tasks, we need to insert tool-use traces into the dialogue to simulate the process of the model invoking the tool API actively.
% First, we reformat the ED dataset into a chatbot format.
% In order to adapt the models to the Tool Learning based Empathetic Response Generation Task
Therefore, we insert tool-use traces into the dialogue to simulate the process of the active tool call, used for training the models.
Although the output of various knowledge bases may not be related, they can achieve a plug-and-play functionality through the interchange of API ports once encapsulated as tools.
When constructing the dataset, we designate COMET as the representative knowledge base tool and default to setting the output of COMET as the execution result of the tool.
In practice, we treat COMET merely as a plugin, so the process can be randomly substituted with other knowledge bases.

Therefore, determining whether to make a tool call is crucial, which requires deep understanding of comprehensive information, including the user's emotional intensity, dialogue context, and so on.
This necessitates the assistance of a LLM with strong capabilities for accurate judgment.
LLaMA3-70B \cite{llama3modelcard} excels in text generation, understanding, and complex problem-solving, so we utilize it to assist in transforming the ED dataset. 
The responses of the assistant in the ED dataset are served as the golden responses in this paper.
The prompt templates are listed in Appendix \ref{app:llama3} and the example of tool usage instance is listed in Appendix \ref{app:casestudy}.
The LLM performs two main tasks:

\textbf{Annotator} aims to determine the appropriateness of invoking the tool based on the context of the conversation, the definition of the empathetic tool, the definition of the annotation task, and the emotion intensity of the users' response.

\textbf{Reflector} is designed to judge the relevance between the result of tool and golden response.
If Annotator determines that the tool needs to be invoked, the Reflector must evaluate the correlation between the results generated by COMET and the responses of the assistant in the ED dataset based on the tool’s definition, causal consistency, intent consistency, and emotional consistency. 
Only if there is a high relevance will the tool usage process be inserted into the \textbf{TOOL-ED} dataset.
In this way, the final dataset will be a chatbot-style dataset that includes tool-use traces.

Based on the above, as for the training dataset, we incorporate tool calls at a rate of 26.46\% and select 10\% of the training set randomly as a validation set.
In order to investigate whether our constructed TOOL-ED enables LLMs to generate higher quality empathetic responses by utilizing the empathetic tools, we conduct a comprehensive evaluation on the test set of the ED dataset.

% Therefore, effectively activating and utilizing this knowledge is crucial for improving task performance.
% Previously, there were two approaches to handling empathetic tasks: the first approach involves fine-tuning the model for specific tasks, while the second uses knowledge graphs directly \cite{qian2023harnessing}. 
% However, both methods will introduce noise and impact performance. 
% To this end, we use commomsense knowledge base as a tool to augment the dialogue context, improving the model generating more appropriate responses.

\subsection{Training Strategy}
% The model-generated utterances (assistant and function\_call) appear in the even-numbered sequences of the dialogue, while the speaker's responses (user) and the tool's output (observation) appear in the odd-numbered sequences.
% For the dialogue response generation, we use the symbol \( \theta \) to represent the dialogue model, with \( \text{observation} \) corresponding to the return result of the tool.
% We opt LoRA-Tuning \cite{hu2021lora} with TOOL-ED dataset to teach the model how to utilize the empathetic tool for better response generation .
Given a dialogue text $U = [u_{1}, u_{2}, ...,u_{n}]$ of length $n$, including roles of user, assistant, function\_call, and observation.
The context of the dialogue consists of utterances from user and assistant, formally represented as $C = [u_{1}, u_{2}, ...,u_{n-1}]$.
For the dialogue response generation, we use the symbol \( \theta \) to represent the dialogue model.
Our objective is to train the model to automatically determine whether it needs to utilize a knowledge base as an auxiliary tool and to better integrate the content returned by the tool with the generation of the model.
We apply LoRA-Tuning \cite{hu2021lora} with models on the TOOL-ED dataset.
% We opt LoRA-Tuning \cite{hu2021lora} with TOOL-ED dataset to teach the model how to utilize the empathetic tool for better response generation .
The objective is to predict the response $u_t$ that will follow the $t-1$ round of dialogue context $C$.
% {
% \begin{equation}
% u_{t} \sim P_{\theta} \left ( \cdot \mid C \right ) 
% \end{equation}
% }
\begin{align}
        \begin{aligned}
         u_{t} \sim P_{\theta} \left ( \cdot \mid C \right ) 
        \end{aligned}
\end{align}
\noindent where $u_t$ has been set with specific formats, including the generation format for tool calls and ordinary text.
And the loss calculation for the tool-based empathetic dialogue task is as follows:
{
\begin{equation}
L_{p} = \sum_{t}^{N} - logP(u_{t} \mid C,\theta)
\end{equation}
}

\begin{table*}
  \centering
  \resizebox{\textwidth}{!}{
  \begin{tabular}{lcccc}
    % \hline
    \toprule
    \textbf{Model}           & \textbf{BLEU-1/2/3/4} & \textbf{B-S}  & \textbf{ROU-1/2/L.} & \textbf{Dist-1/2} \\
    \hline
     CEM*              & 0.1332/0.0630/0.0351/0.0209         & 0.8603        & 0.1625/0.0418/0.1508   & 0.0066/0.0299 \\
    EmpDG          & 0.1857/0.0873/0.051/0.0316       & 0.8611        & 0.1697/0.0453/0.155   & 0.0181/0.0694 \\
     HEF              & 0.1164/0.0365/0.0171/0.0084         & 0.8539        & 0.1471/0.0186/0.1196   & 0.0336/\textbf{0.2096} \\
    KEMP*       & 0.1902/0.0767/0.0362/0.0195   & 0.8531        & 0.1618/000296/0.1418   & 0.0041/0.0204 \\
    IAMM       & 0.1505/0.0651/0.0360/0.0215   & 0.8633        & 0.1580/0.0347/0.1446   & 0.0098/0.0302 \\
    MOEL       & 0.1726/0.0773/0.0433/0.0264   & 0.8582        & 0.1653/0.0367/0.1502   & 0.0038/0.0160 \\
    MIME       & 0.2182/0.096/0.0492/0.02796   & 0.8613        & 0.1882/0.0387/0.1642   & 0.0032/0.0124 \\
    \hline
    vicuna\_base & 0.1106/0.0412/0.0209/0.0111   & 0.8522        & 0.1525/0.0271/0.1254   & 0.0274/0.1833 \\
    vicuna\_oneshot & 0.1204/0.0449/0.0227/0.012   & 0.8544        & 0.1550/0.0259/0.1273   & 0.0283/0.1793 \\
    vicuna\_lora & \textbf{0.1935}/0.0879/0.0514/0.0316   & 0.8723        & 0.1740/0.0403/0.1561   & 0.0296/0.1424 \\
    \hdashline
    vicuna\_tool\_comet* & 0.1859/0.0903/0.0544/0.0339   & \textbf{0.8755}        & 0.1888/\textbf{0.0543}/\textbf{0.1735}   & 0.0299/0.1453 \\
    vicuna\_tool\_cicero* & 0.1877/\textbf{0.0904}/\textbf{0.0545}/\textbf{0.0341}   & 0.8751        &\textbf{ 0.1889}/0.0534/0.1732   & \textbf{0.0300}/0.1447 \\
    \hline
    % vicuna\_cot & 0.0805/0.0272/0.0118/0.0052   & 0.8363        & 0.1204/0.0148/0.0952   & 0.0182/0.1292 \\
    qwen\_base & 0.1046/0.0349/0.016/0.0077   & 0.8456        & 0.1356/0.0186/0.1072   & \textbf{0.2322}/0.1720 \\
    qwen\_oneshot & 0.1158/0.0253/0.1780/0.0078   & 0.8497        & 0.1431/0.0211/0.1158   & 0.0281/0.1904 \\
    qwen\_lora & 0.1861/0.0883/0.0517/0.0317   & 0.8760        & 0.1846/0.0483/0.1696   & 0.0249/0.1172 \\
    % qwen14b\_cot & 0.0805/0.0272/0.0118/0.0052   & 0.8363        & 0.1204/0.0148/0.0952   & 0.0182/0.1292 \\
    \hdashline
    % vicuna\_oneshot & 0.0631/0.0244/0.0122/0.0062   & 0.8351        & 0.1358/0.0171/0.1102   & 0.0281/\textbf{0.1904} \\    
    % vicuna\_lora & \textbf{0.1935}/0.0879/0.0514/0.0316   & 0.8723        & 0.174/0.0403/0.1561   & 0.0296/0.1424 \\   
    % qwen\_oneshot & 0.1158/0.0253/0.178/0.0078   & 0.8497        & 0.1431/0.0211/0.1158   & 0.0281/\textbf{0.1904} \\    
    % qwen\_lora & 0.1861/0.883/0.0517/0.0317   & 0.876        & 0.1846/0.0483/0.1696   & 0.0249/0.1172 \\    
    qwen\_tool\_comet* & \textbf{0.1936}/
\textbf{0.0961}/\textbf{0.0579}/\textbf{0.036}   & 0.8765        & \textbf{0.1941}/\textbf{0.0579}/\textbf{0.1793 }  & 0.0283/0.1358 \\
    qwen\_tool\_cicero* & 0.1907/0.0943/0.0569/0.0352   & \textbf{0.8766}        & 0.1913/0.0563/0.1768   & \textbf{0.0289}/\textbf{0.1395} \\
    % \hline
    \bottomrule
  \end{tabular}
  }
  \caption{\label{table-1}
    Results of automic evaluation between EKTC models and baselines. (i) "*" denotes the models that integrate external knowledge. (ii) vicuna\_base and qwen\_base separately represent the base models of Vicuna-7B and Qwen1.5-14B.
(iii) vicuna\_lora and qwen\_lora respectively refer to the base models fine-tuned on the original ED dataset. 
(iv) vicuna\_tool\_comet and qwen\_tool\_comet denote the results of the COMET as tool after the base models fine-tuned on the TOOL-ED dataset.
(v) vicuna\_tool\_cicero and qwen\_tool\_cicero signify the result of utilizing CICERO as tool after the corresponding base models fine-tuned on the TOOL-ED dataset.
  }
\end{table*}

\section{Experiment}

\subsection{Dataset}
The ED dataset includes 32 emotion labels and corresponding contexts for each dialogue, which contains 33,090 dialogues.
% In these dialogues, speakers discuss their situations, and listeners try to understand the speakers' emotions and provide appropriate responses. 
Based on the ED dataset, we have added tool usage traces to reconstruct the TOOL-ED dataset.

\subsection{Baselines}
To verify the effectiveness of our EKTC framework, we choose the following state-of-the-art (SOTA) models, and conduct a comparative evaluation based on their test results on the ED dataset:

\textbf{CEM} \cite{sabour2022cem} uses commonsense knowledge to enhance the understanding of the conversational context and the feelings of the interlocutor. 
\textbf{EmpDG} \cite{li-etal-2020-empdg} combines dialogue-level and word-level sentiment analysis with an interactive adversarial learning framework to more precisely capture user emotions and generate high-quality responses.
\textbf{HEF} \cite{yang2024enhancing} combines LLMs with  small models, using a two-stage emotion prediction strategy to help the LLM focus on the primary emotions emphasized by the smaller model.
\textbf{KEMP} \cite{li2022knowledge} utilizes external knowledge graphs to extract emotional signals, learning emotional dependencies through an emotional cross-attention mechanism.
\textbf{IAMM} \cite{yang2024iterative} employs a second-order interactive attention mechanism to enhance the understanding capability of dialogue systems by capturing important associative words in conversations.
\textbf{MOEL} \cite{lin-etal-2019-moel} is an innovative end-to-end empathy modeling approach that captures user emotions and outputs an emotional distribution.
\textbf{MIME} \cite{majumder-etal-2020-mime} groups emotions based on their positivity or negativity, with responses mimicking the user's emotions to varying degrees, enhancing empathy and contextual relevance in the responses.
We replicate them based on the open-source code of the project and conduct testing experiments on the ED dataset.

% As for LLM baselines, Llama3 \cite{meta2024introducing} is the latest generation of artificial intelligence language models developed by Meta, with scales of 8B and 70B.
% ChatGLM-3 \cite{glm2024chatglm, du2021glm, zeng2022glm} is the generation language model jointly developed by Chinese companies Tsinghua University and Zhipu AI.
% The Mistral7B \cite{jiang2023mistral, jung2010mistral} is a language model with 7 billion parameters released by Mistral AI.
% As for LLM baselines, we benchmark against Vicuna-7B \cite{chiang2023vicuna} and Qwen-14B \cite{bai2023qwen} to fine-tune them respectively on the ED dataset which is formatted as a dialogue and our TOOL-ED dataset. 

As for LLM baselines, we benchmark against Vicuna-7B \cite{chiang2023vicuna} and Qwen1.5-14B \cite{bai2023qwen} to fine-tune them on the TOOL-ED dataset and the ED dataset for comparision.

\subsection{Implementation Details}

We fine-tune Vicuna-7B \cite{chiang2023vicuna} and Qwen1.5-14B  \cite{bai2023qwen} as baselines on the TOOL-ED corpus.
To demonstrate the effectiveness of the EKTC framework, we also fine-tuned the model on the ED dataset formatted in dialogue format for comparison.
% We fine-tune the LLMs on the TOOL\_EMAPTHETICDIALOGUE corpus. 
% For the LLMs, we benchmark against Vicuna-7B \cite{chiang2023vicuna} and Qwen1.5-14B \cite{bai2023qwen} models.
% The fine-tuned models can utilize the defined commonsense Knowledge Base tool API to assist in response generation. 
The training procedure is on NVIDIA A6000 48G GPUs using the LLaMA-Factory framework\footnote{\href{https://github.com/hiyouga/ LLaMA-Factory}{https://github.com/hiyouga/ LLaMA-Factory}}.
% The ratio of utilizing COMET and CICERO knowledge bases in this papaer is shown in the Figure~\ref{figure_12}.
Moreover, we conduct oneshot tests on Vicuna-7B and Qwen1.5-14B on the ED dataset separately. 
% We guide LLaMA3 to assist injecting external knowledge into the conversation based on the tool call timing it has marked.
We divide the Tool-ED dataset and the ED dataset into training, validation, testing with 8:1:1 ratio referring to \cite{rashkin2018towards}. The parameter settings of all SOTA baseline models are consistent with those recommended in their initial paper or code.
As for EKTC framework, we employ the COMET of BART version \cite{hwang2021comet} and CICERO \cite{shen2022multiview} as the empathetic tools to evaluate the results on the ED dataset.

\begin{table}[h]
\centering
\resizebox{\columnwidth}{!}{
\begin{tabular}{p{2.2cm}cccc}
\hline
Comparisons & Aspects & Win & Tie & Lose \\
\hline
\multirow{4}{*}{\parbox[t]{2.0cm}{qwen\_tool\_comet \\ vs. qwen\_base}} &  Emp. & 64.0\%  & 24.0\% & 12.0\% \\
& Inf. & 51.3\% & 40.7\% & 8.0\% \\
& Flu. & 17.0\% & 71.3\% & 11.7\% \\
& Con. & 67.7\% & 24.3\% & 8.0\%\\
\hline
\multirow{4}{*}{\parbox[t]{2.0cm}{qwen\_tool\_comet \\ vs. qwen\_lora}} &  Emp. & 61.3\%  & 32.7\% & 6.0\% \\
& Inf. & 52.0\% & 35.3\% & 12.7\% \\
& Flu. & 60.0\% & 32.3\% & 7.7\% \\
& Con. & 63.7\% & 24.0\% & 12.3\%\\
\hline
\multirow{4}{*}{\parbox[t]{2.0cm}{vicuna\_tool\_comet \\ vs. \\vicuna\_base}} &  Emp. & 57.7\%  & 28.0\% & 14.33\% \\
& Inf. & 15.7\% & 45.0\% & 39.3\% \\
& Flu. & 34.7\% & 42.3\% & 23.0\% \\
& Con. & 61.0\% & 26.0\% & 13.0\%\\
\hline
\multirow{4}{*}{\parbox[t]{2.0cm}{vicuna\_tool\_comet \\ vs. \\vicuna\_lora}} &  Emp. & 72.3\%  & 22.3\% & 5.3\% \\
& Inf. & 53.3\% & 33.3\% & 13.3\% \\
& Flu. & 67.0\% & 29.7\% & 3.3\% \\
& Con. & 67.7\% & 27.0\% & 5.33\%\\
\hline
\end{tabular}
}
\caption{\label{table-2}
Results of human evaluation on aspects.
 }
\end{table}

\begin{table}[h]
\centering

\begin{tabular}{p{2.5cm}ccc}
\hline
Comparisons & Aspects & Win & Lose \\
\hline

\multirow{3}{2.5cm}{\centering qwen\_tool\_comet \\ vs. qwen\_lora} &  Emp. & 71\%  & 29\% \\
& Con. & 86\% & 14\% \\
& Flu. & 87\% & 13\%\\
\hline
\multirow{3}{2.5cm}{\centering vicuna\_tool\_comet \\ vs. vicuna\_lora} &  Emp. & 77\%  & 23\% \\
& Con. & 83\% & 17\%\\
& Flu. & 90\% & 10\%\\
\hline
\end{tabular}

\caption{\label{table-5}
Results of LLM-based evaluation on aspects.
 }
\end{table}

\begin{table*}
  \centering
  \resizebox{\textwidth}{!}
 {
  \begin{tabular}{lcccc}
    % \hline
    \toprule
    \textbf{Model}           & \textbf{BLEU-1/2/3/4} & \textbf{B-S}  & \textbf{ROU-1/2/L} & \textbf{Dist-1/2} \\
    \hline
    qwen\_kno          & 0.1105/0.0377/0.0178/0.0086       & 0.8473        & 0.1413/0.021/0.1143   & 0.0294/\textbf{0.2114 }\\
    qwen\_noref\_comet         & \textbf{0.1971}/0.0919/0.0527/0.3130      & 0.8722        & 0.1797/0.0471/0.1654   & 0.0298/0.1409 \\
     qwen\_tool\_comet             & 0.1936/\textbf{0.0961}/\textbf{0.0579}/\textbf{0.0360}         & 0.8765        & \textbf{0.1941}/\textbf{0.0579}/\textbf{0.1793}   & 0.0283/0.1358 \\
     qwen\_noref\_cicero         & 0.1927/0.0910/0.0528/0.3190      & 0.8740        & 0.1828/0.0499/0.1683   & \textbf{0.0316}/0.1464 \\
     qwen\_tool\_cicero             & 0.1907/0.0943/0.0569/0.0352         & \textbf{0.8766 }       & 0.1913/0.0563/0.1768   & 0.0289/0.1395 \\
    \hline
    vicuna\_kno       & 0.0905/0.0299/0.0139/0.0069   & 0.8455        & 0.1156/0.0155/0.0974   & \textbf{0.0358}/\textbf{0.2188} \\
    vicuna\_noref\_comet          & \textbf{0.1943}/\textbf{0.0925}/\textbf{0.055}/\textbf{0.0341}       & 0.8730        & 0.1859/0.0527/0.1701   & 0.0290/0.1418 \\
    vicuna\_tool\_comet       & 0.1859/0.0903/0.0544/0.0339   & \textbf{0.8755 }       & 0.1888/\textbf{0.0543}/\textbf{0.1735}   & 0.0299/0.1453 \\
    vicuna\_noref\_cicero          & 0.1586/0.0731/0.0419/0.0248       & 0.8638        & 0.1845/0.0501/0.1656   & 0.0283/0.1427 \\
    vicuna\_tool\_cicero       & 0.1877/0.0904/0.0545/0.0341   & 0.8751       & \textbf{0.1889}/0.0534/0.1731   & 0.0299/0.1453 \\
    % \hline
    \bottomrule

  \end{tabular}
 }
  \caption{\label{table-3}
 Results of evaluation for ablation study.
  }
\end{table*}

\subsection{Evaluation Metrics}
% In order to verify the effectiveness of EKTC, we employ comprehensive evaluations and select abundant metrics:
% \\
\textbf{Automatic Evaluation}
We utilize Distinct-n (Dist-1/2) \cite{li-etal-2016-diversity}, BERTScore (B-S) \cite{zhang2019bertscore}, ROUGE (ROU-1/2/L.) \cite{fang2023eva} and BLEU-n (BLEU-1/2/3/4) \cite{papineni2002bleu} as the primary automatic metrics for evaluating response empathetic generation performance.
Distinct-1 and Distinct-2 assess response diversity at the unigram and bigram levels respectively.
B-S leverages the pre-trained embeddings of BERT and matches words in candidate sentences with those in reference sentences on cosine similarity.
ROUGE and BLEU-n measures the similarity and relevance between generated responses and reference responses.
\\
\textbf{Human Evaluation}
Human evaluation remains essential for a thorough and nuanced understanding of content quality and effectiveness.
Following previous methods \cite{sabour2022cem}, we use A/B testing to compare the baseline models with our model. 
We randomly select 100 conversation samples and compare the performance of the baseline model with the qwen\_tool model in pairs.
We recruited three researchers specializing in emotional dialogue systems as annotators, excluding the authors of this paper. 
We evaluate from four aspects: 
\textbf{Empathy} (Emp.) measures whether the emotional response sufficiently understands the users' emotions and intentions and generates an appropriate reply.
\textbf{Informativity} (Inf.) meatures whether the response contains valuable information.
\textbf{Fluency} (Flu.) measures whether the response is similar to human expression, natural, and smooth.
\textbf{Consistency} (Con.) measures whether the response is concise, clear, and relevant to the topic.
For the same dialogue, if our model performs better, it is annotated as Win. If it performs worse, it is annotated as Lose. If there is little difference between the two, it is annotated as Tie.

\textbf{LLM-based Evaluation}
GPT-4 achieves a high degree of similarity to human evaluations, so we opt it to simulate human assessors for evaluating the performance of other models.
We continue to assess three aspects including \textbf{Empathy}, \textbf{Fluency}, and \textbf{Consistency} by conducting an A/B test between the models fine-tuned with LoRA-Tuning on the original ED dataset and the models fine-tuned on our TOOL-ED dataset with the COMET knowledge base as the tool.

\subsection{Results and Analysis}
\subsubsection{Main Results}
\textbf{Automatic Evaluation}
Table~\ref{table-1} shows the automatic evaluation results.
The EKTC-based models invoking the empathetic tools outperform most compared models in terms of the evaluation metrics, demonstrating the superior comprehension and expression capabilities of the EKTC-based models. 
When utilizing the two different empathetic tools (COMET and CICERO), we simply replacing the API ports of the tools, observing that the performance of the model remains extremely stable.
This indicates that  tool learning in this end-to-end mode allows the model to operate independently of any specific knowledge base, demonstrating the robustness and generalizability of our framework.
In terms of Dist-n, the performances of the EKTC-based models are slightly lower than some of the LLMs.
This could be attributed to the fact that the results returned by the the knowledge base are also incorporated into the dialogue history, which limits the length of conversation history input into the model. However, a slight reduction in the diversity of empathetic responses may not necessarily be a negative outcome.
\\
\textbf{Human Evaluation}
Table~\ref{table-2} shows the results of human evaluations. 
Compared with the baseline fine-tuned directly on the ED dataset, the EKTC-based models demonstrates superior empathy capabilities, which is primarily due to the reasonable utilization of knowledge base tools, assisting the models better understand the speaker's emotions. 
The advantage in relevance and fluency is attributed to the EKTC paradigm facilitating the model's ability to navigate emotional expression and phrasing.
\\
\textbf{LLM-based Evaluation}
Table~\ref{table-5} shows the results of LLM-based evaluations. The EKTC-based models exhibit better empathy than the baseline fine-tuned on the ED dataset, mainly due to the appropriate use of knowledge base tools that assist the model in getting better understanding of the emotions of users. 
The advantage in relevance and 
 fluency primarily stems from the reasonable allocation of the knowledge base tools improves the models' cognition of the users' intentions.
% The strength in fluency attributes to the EKTC paradigm can facilitate model's processing of emotional expression and phrasing.

% \begin{table}[h]
% \centering

% \begin{tabular}{p{2.5cm}ccc}
% \hline
% Comparisons & Aspects & Win & Lose \\
% \hline

% \multirow{3}{2.5cm}{\centering qwen\_tool\_v1 \\ vs. qwen\_lora} &  Emp. & 71\%  & 29\% \\
% & Con. & 86\% & 14\% \\
% & Flu. & 87\% & 13\%\\
% \hline
% \multirow{3}{2.5cm}{\centering vicuna\_tool\_v1 \\ vs. vicuna\_lora} &  Emp. & 77\%  & 23\% \\
% & Con. & 83\% & 17\%\\
% & Flu. & 90\% & 10\%\\
% \hline
% \end{tabular}

% \caption{\label{table-5}
% Results of LLM-based evaluation on aspects.
%  }
% \end{table}

\begin{figure}[h]
\centering
\includegraphics[width=0.5\textwidth]{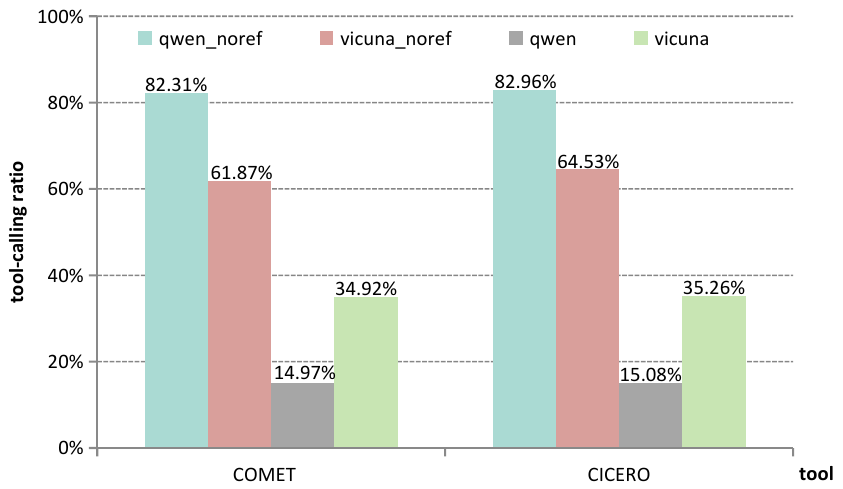}
\caption{Tool-calling ratio of the fine-tuned models.
(i) qwen\_noref and vicuna\_noref refer to Qwen1.5-14B and Vicuna-7B models after fine-tuned without reflection processes for ablation experiments. (ii) qwen and vicuna refer to the results of fine-tuned Qwen1.5-14B and Vicuna-7B on the TOOL-ED dataset.
}
% (i) qwen\_noref and vicuna\_noref refer to Qwen1.5-14B and Vicuna-7B models after fine-tuned without reflection processes for ablation experiments. (ii) qwen and vicuna refer to the results of fine-tuned Qwen1.5-14B and Vicuna-7B on the TOOL-ED dataset.}
\label{figure_12}
\end{figure}

\subsubsection{Ablation Studies}
We constructed the following ablation models for comparison: 

(i) models using COMET in each round of dialogue with appropriate prompt for relevant knowledge reasoning, represented as qwen\_kno and vicuna\_kno.

(ii) models fine-tuned on the dataset where judge the tool calling process without adding a reflection process, represented as qwen\_noref\_comet, vicuna\_noref\_comet, qwen\_noref\_cicero and vicuna\_noref\_cicero.

% \noindent(iii) the EKTC-based models, represented as qwen\_tool\_comet, vicuna\_tool\_comet, qwen\_tool\_cicero and vicuna\_tool\_cicero.

(iii) The EKTC-based models, fine-tuned on the TOOL-ED, represented as qwen\_tool\_comet, vicuna\_tool\_comet, qwen\_tool\_cicero, vicuna\_t-
ool\_cicero.

Table~\ref{table-3} and Figure~\ref{figure_12} respectively present the results of the ablation experiments and the tool invocation ratios of the models. Although the EKTC-based models has a relatively lower tool calling ratio, the performance in generating empathic dialogue is actually better.
This suggests that the timing of tool invocation designed under the EKTC framework effectively improves the quality of the responses of the models.
By comparing the response generation effects of models that utilize the knowledgebase in each round with the EKTC-based models, we can prove that our framework effectively mitigates the impact of noise.

\section{Conclusion}
In this paper, we propose EKTC, which is a tool-based empathetic dialogue paradigm. 
To enable more models to be adapted to the task, we reconstruct the TOOL-ED dataset based on the ED dataset.
Then we define the knowledge bases as tools, which efficiently stimulates relevant knowledge encoded by LLM and avoid noise from external knowledge in an end-to-end way. 
We fine-tune the models on the reconstructed dataset and validate the effectiveness of our approach through both automatic and human evaluations.
Furthermore, by replacing various knowledge bases tools in a plug-and-play manner to test the ability of the models to generate empathetic responses, we demonstrate the generalizability of EKTC.
In the future, our proposed framework can integrate more tools and be applied to a wider range of downstream tasks.

\section*{Limitation}
We has certain limitations in its definition of tools, as it does not cover the process of using multiple tools. 
Defining a broader range of tools and incorporating external knowledge through a hybrid approach could potentially further enhance the richness of empathetic dialogue generation.

\section*{Acknowledgements}

This work is supported by the National Natural Science Foundation of China (No. 62272092, No. 62172086) and the Fundamental Research Funds for the Central Universities of China (No. N2116008).

% \section*{Acknowledgments}

% This document has been adapted by Emily Allaway from the instructions for earlier ACL and NAACL proceedings, including those for NAACL 2024 by Steven Bethard, Ryan Cotterell and Rui Yan,
% ACL 2019 by Douwe Kiela and Ivan Vuli\'{c},
% NAACL 2019 by Stephanie Lukin and Alla Roskovskaya,
% ACL 2018 by Shay Cohen, Kevin Gimpel, and Wei Lu,
% NAACL 2018 by Margaret Mitchell and Stephanie Lukin,
% Bib\TeX{} suggestions for (NA)ACL 2017/2018 from Jason Eisner,
% ACL 2017 by Dan Gildea and Min-Yen Kan,
% NAACL 2017 by Margaret Mitchell,
% ACL 2012 by Maggie Li and Michael White,
% ACL 2010 by Jing-Shin Chang and Philipp Koehn,
% ACL 2008 by Johanna D. Moore, Simone Teufel, James Allan, and Sadaoki Furui,
% ACL 2005 by Hwee Tou Ng and Kemal Oflazer,
% ACL 2002 by Eugene Charniak and Dekang Lin,
% and earlier ACL and EACL formats written by several people, including
% John Chen, Henry S. Thompson and Donald Walker.
% Additional elements were taken from the formatting instructions of the \emph{International Joint Conference on Artificial Intelligence} and the \emph{Conference on Computer Vision and Pattern Recognition}.

\bibliography{custom}

\clearpage
\appendix

\section{Prompt}
\label{sec:appendix}
In this section, we will present detailed information about the prompt templates in our work. 
\subsection{Prompt for Anotator and Reflector}
\begin{figure*}
\centering
\includegraphics[width=0.8\textwidth]{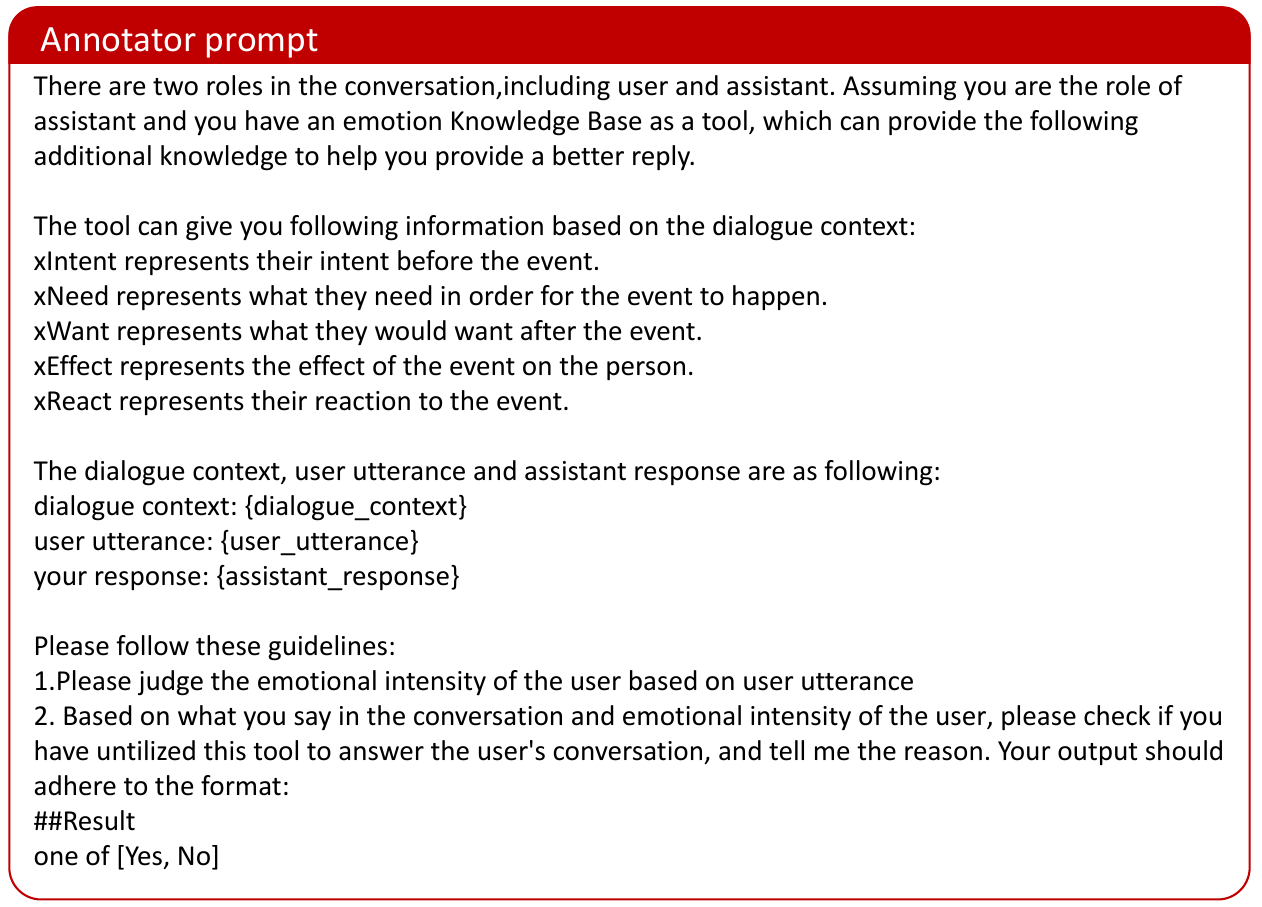}
\caption{Prompt for Anotator}
\label{figure_5}
\end{figure*}

% Figure~\ref{figure_5} and Figure~\ref{figure_6} show the prompt of data construction, including prompt for annotator and reflector.
Figure~\ref{figure_5} show the prompt for \textbf{Annotator} during data construction.
First, we provide a comprehensive definition of the empathy dialogue task and the annotation task. 
Next, we offer a detailed explanation of the defination and the result of the $Emotionknowledgebase$ tool. 
Finally, we supply annotator with contextual information and the gold responses from the ED dataset, we instruct them to evaluate whether the current conversation state represents the optimal moment for tool invocation.

\begin{figure*}
\centering
\includegraphics[width=0.8\textwidth]{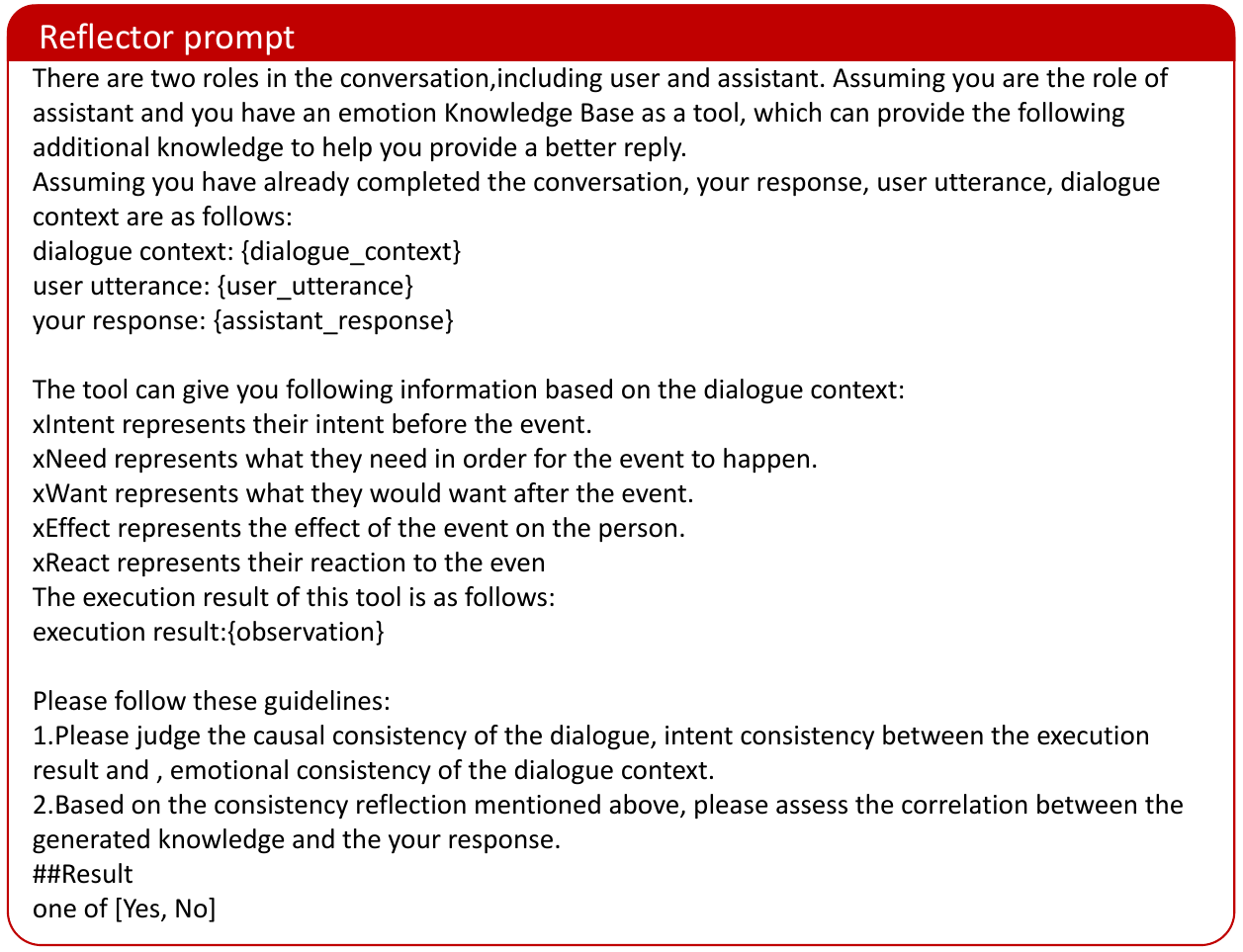}
\caption{Prompt for Reflector}
\label{figure_6}
\end{figure*}

Figure~\ref{figure_6} show the prompt for \textbf{Reflector} during data construction.
Similar to the prompts provided for annotators, we specify detailed definitions for the empathy dialogue task, relevance judgment task, and tool definitions. 
We instruct Reflector to assess the relevance between the output of the tool and the gold responses from the ED dataset, based on causal consistency, intent consistency, and emotional consistency.

\label{app:llama3}

\subsection{Prompt Template for Tool Learning.}
\begin{figure*}
\centering
\includegraphics[width=0.8\textwidth]{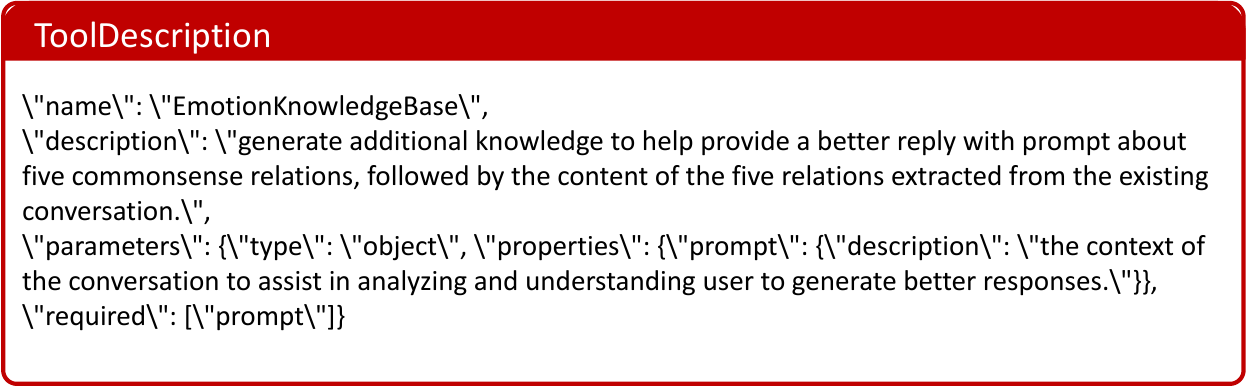}
\caption{Discription of the empathetic tool}
\label{figure_13}
\end{figure*}

\begin{figure*}
\centering
\includegraphics[width=0.8\textwidth]{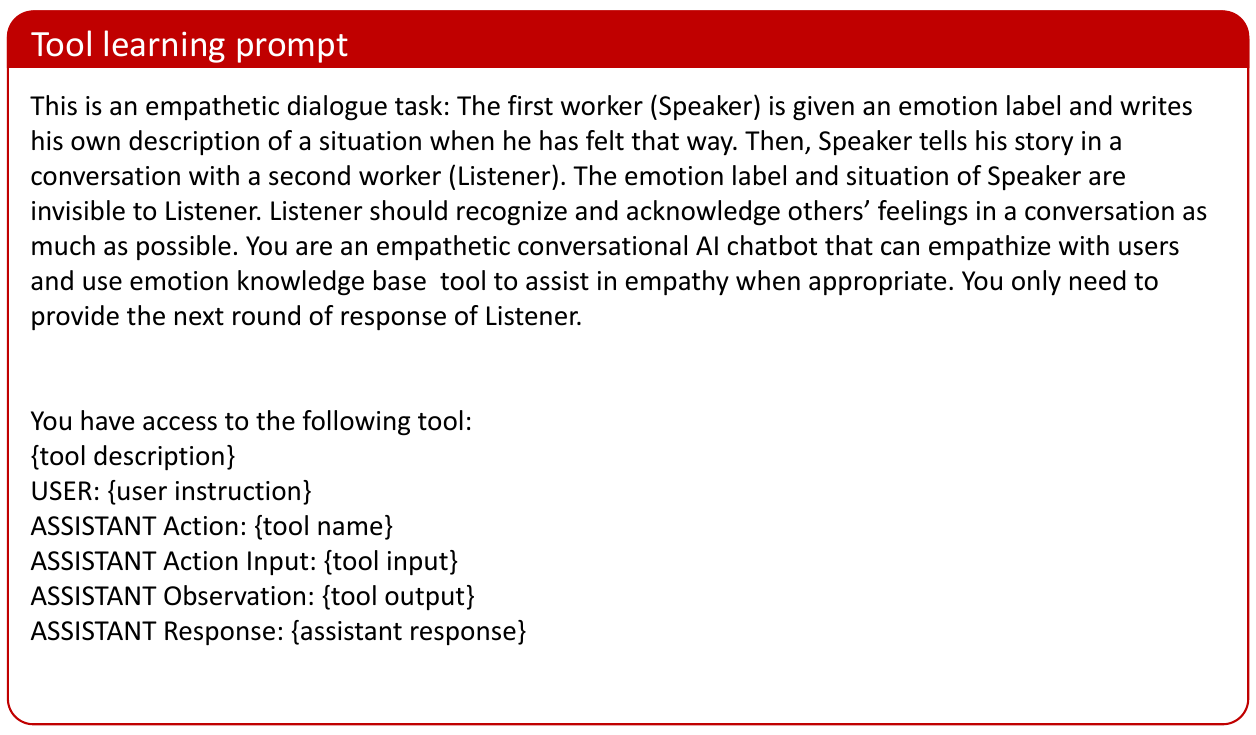}
\caption{Prompt for Tool learning}
\label{figure_7}
\end{figure*}

Based on \cite{zhang-etal-2024-stickerconv} prompt about tool learning, we have modified the format according to the characteristics of empathy dialogue tasks. The specific details of the prompt are shown in the Figure~\ref{figure_7}. Figure~\ref{figure_13} presents the discription of the empathetic tool.

\label{app:toollearning}

\subsection{Prompt Template for Tool LLM-based evaluation.}
\begin{figure*}
\centering
\includegraphics[width=0.8\textwidth]{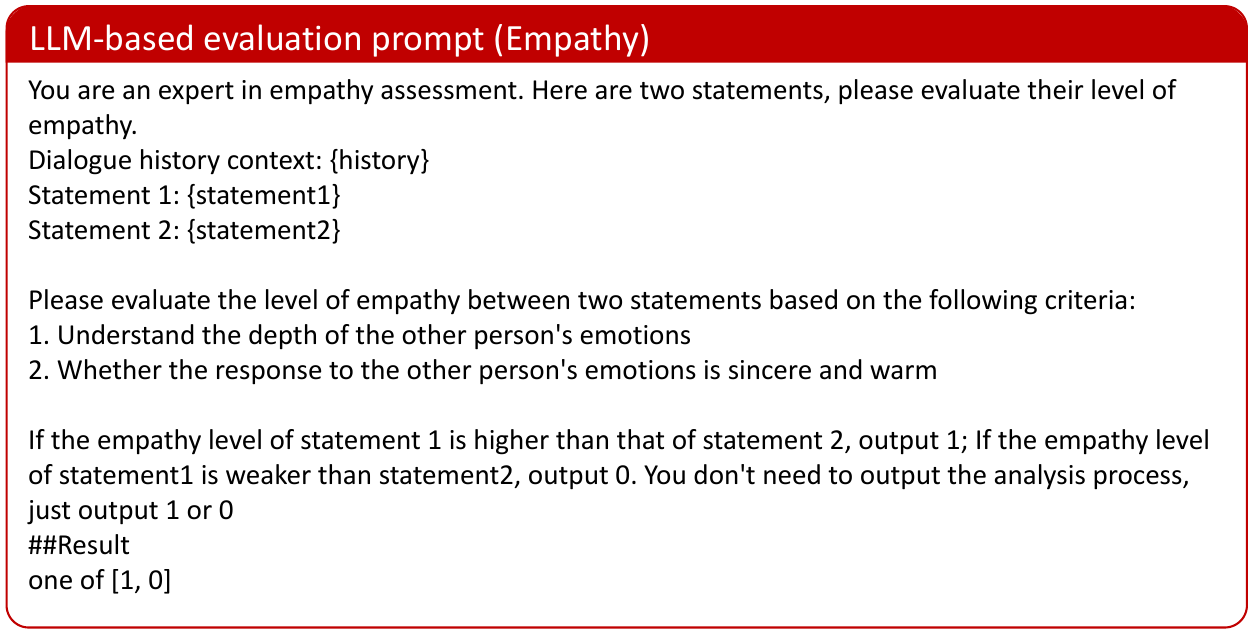}
\caption{Prompt Template for Empathy Scorer}
\label{figure_9}
\end{figure*}

\begin{figure*}
\centering
\includegraphics[width=0.8\textwidth]{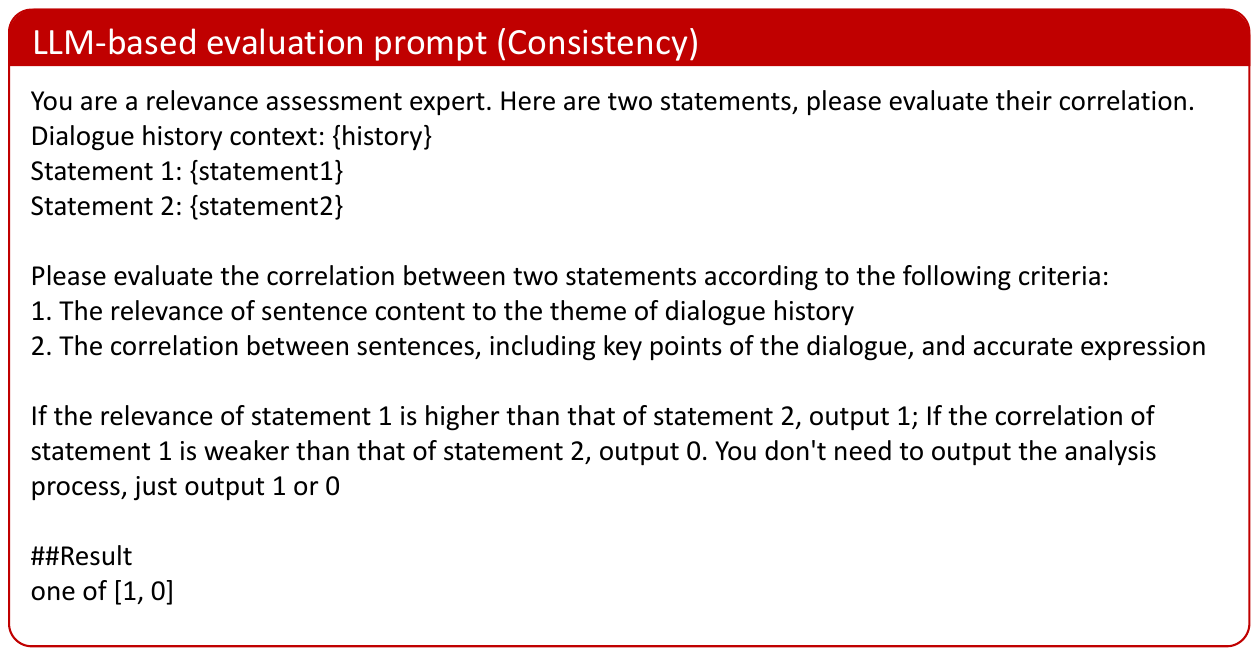}
\caption{Prompt Template for Consistency Scorer}
\label{figure_10}
\end{figure*}

\begin{figure*}
\centering
\includegraphics[width=0.8\textwidth]{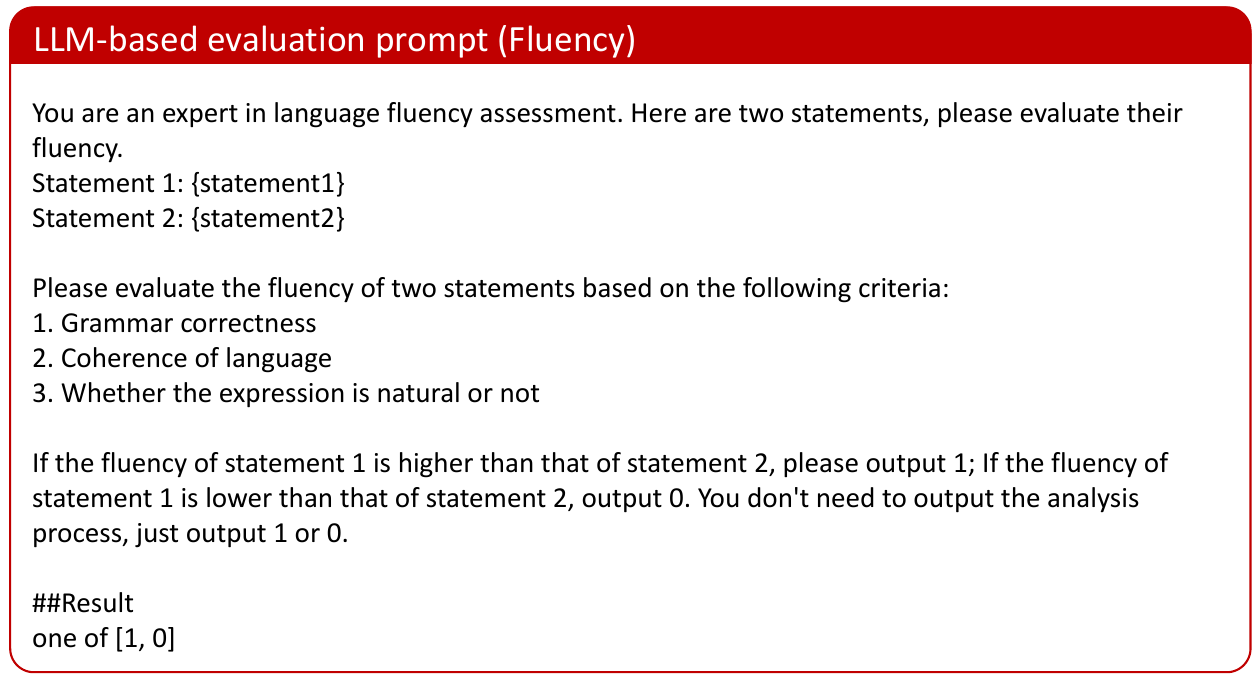}
\caption{Prompt Template for Fluency Scorer}
\label{figure_11}
\end{figure*}

We take advantage of gpt4 to implement LLM evaluation
The prompt of specific empathy, consistency, and fluency evaluation metric are shown in the Figure~\ref{figure_9}, Figure~\ref{figure_10}, Figure~\ref{figure_11}.

\label{app:llmbasedeval}

% \section{Example of Tool-Use Instance}
% \label{app:toolusage}
% Figure~\ref{figure_8} show an example of tool-use trajectory in the TOOL-ED dataset.

\section{Case Study}
\label{app:casestudy}
% Figure~\ref{figure_8} show an example of tool-use trajectory in the TOOL-ED dataset.

\begin{figure*}
\centering
\includegraphics[width=0.8\textwidth]{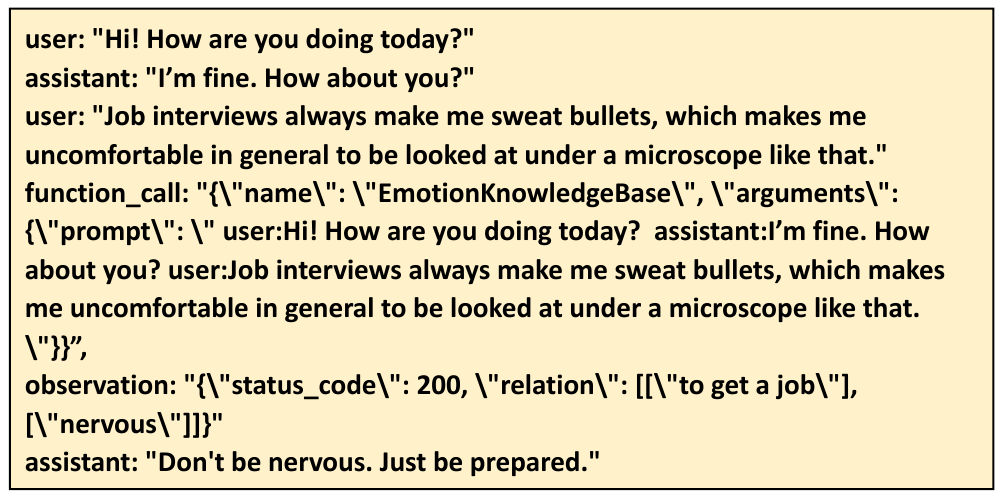}
\caption{Example from the TOOL-ED dataset}
\label{figure_8}
\end{figure*}

To illustrate the tool usage process within the TOOL-ED dataset more clearly, Figure~\ref{figure_8} provides an example of tool-use trajectory in the dataset. 
After the user initiates a conversation, the model can take one of two different response approaches:
(1) Directly generate a reply, such as the example shown in the figure, “I’m fine. How about you?”;
(2) Employ the tool, generating content in JSON format corresponding to function\_call, which includes the name of the tool and the relevant input parameters.
After inputting these parameters into the knowledge base tool, the tool will output relevant commomsense knowledge as the observation through a data flow approach. 
The model then generates a response that is more contextually appropriate and aligned with the conversational tone based on the external knowledge introduced.

\begin{figure}
\centering
\includegraphics[width=0.4\textwidth]{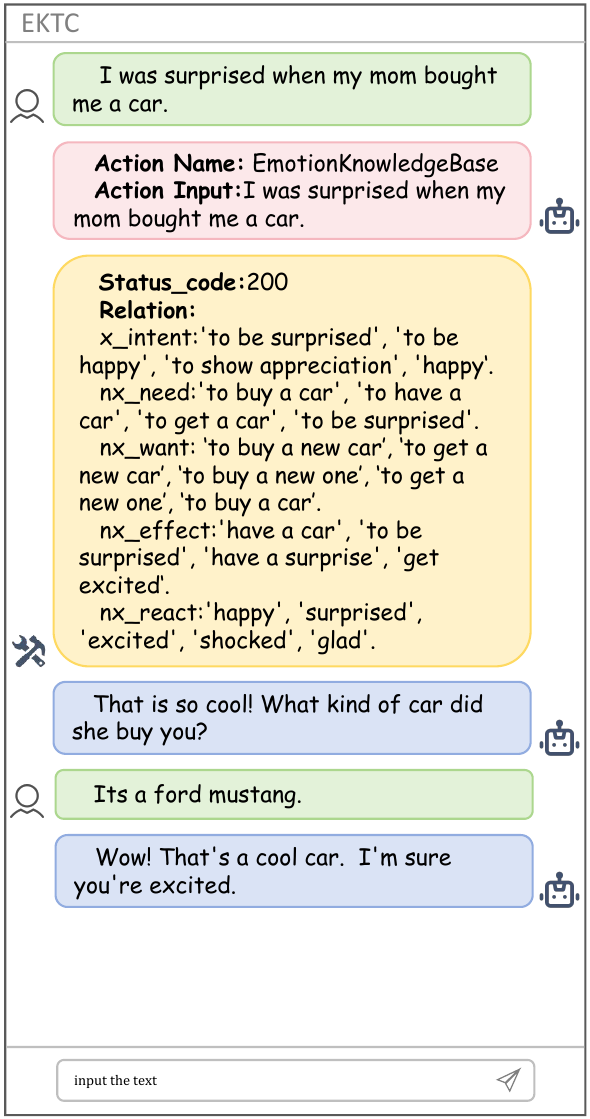}
\caption{Examples of conversations by the user interacting with the EKTC-based model.}
\label{figure_14}
\end{figure}

As shown in Figure~\ref{figure_14}, this case demonstrates the EKTC-based model's ability to actively call the $EmotionKnowledgebase$ tool. By autonomously generating Action names and corresponding Action Inputs, the system can identify the need for tool invocation and then transmits external knowledge through the knowledge base, integrating it into the conversation history. The model subsequently generates a response based on the updated conversation history.


\end{document}